\documentclass[conference]{IEEEtran}
\IEEEoverridecommandlockouts
\usepackage{cite}
\usepackage{amsmath,amssymb,amsfonts}
\usepackage{algorithmic}
\usepackage{graphicx}
\usepackage{textcomp}
\usepackage{xcolor}

\usepackage{graphicx}
\usepackage[numbers]{natbib}
\usepackage{doi}
\usepackage{listings}

\usepackage{listings}
\usepackage{xcolor}

\usepackage{tabularx}
\usepackage{enumitem}
\usepackage{hyperref}

\definecolor{codegreen}{rgb}{0,0.6,0}
\definecolor{codegray}{rgb}{0.5,0.5,0.5}
\definecolor{codepurple}{rgb}{0.58,0,0.82}
\definecolor{backcolour}{rgb}{0.95,0.95,0.92}

\lstdefinestyle{mystyle}{
    backgroundcolor=\color{backcolour},   
    commentstyle=\color{codegreen},
    keywordstyle=\color{magenta},
    numberstyle=\tiny\color{codegray},
    stringstyle=\color{codepurple},
    basicstyle=\ttfamily\footnotesize,
    breakatwhitespace=false,         
    breaklines=true,                 
    captionpos=b,                    
    keepspaces=true,                 
    numbers=left,                    
    numbersep=5pt,                  
    showspaces=false,                
    showstringspaces=false,
    showtabs=false,                  
    tabsize=2
}

\def\BibTeX{{\rm B\kern-.05em{\sc i\kern-.025em b}\kern-.08em
    T\kern-.1667em\lower.7ex\hbox{E}\kern-.125emX}}
\begin{document}

\title{Breaking SafetyCore: Exploring the Risks of On-Device AI Deployment}

\author{\IEEEauthorblockN{Guyomard Victor}
\IEEEauthorblockA{\textit{Skyld AI} \\
Rennes, France \\
victor.guyomard[at]skyld.io}
\and
\IEEEauthorblockN{Mathis Mauvisseau}
\IEEEauthorblockA{\textit{Skyld AI} \\
Rennes, France \\
mathis.mauvisseau[at]skyld.io}
\and
\IEEEauthorblockN{Paindavoine Marie}
\IEEEauthorblockA{\textit{Skyld AI} \\
Rennes, France \\
marie[at]skyld.io}
}

\maketitle

\begin{abstract}
Due to hardware and software improvements, an increasing number of AI models are deployed on-device. This shift enhances privacy and reduces latency, but also introduces security risks distinct from traditional software. In this article, we examine these risks through the real-world case study of SafetyCore, an Android system service incorporating sensitive image content detection. We demonstrate how the on-device AI model can be extracted and manipulated to bypass detection, effectively rendering the protection ineffective. Our analysis exposes vulnerabilities of on-device AI models and provides a practical demonstration of how adversaries can exploit them.
\end{abstract}


\section{Introduction}
Today, an increasing number of AI\footnote{In this article, we use the term AI to specifically refer to deep learning neural networks.} models are deployed on-device. This trend is driven by the growing computational power of modern hardware, especially with the adoption of specialized components like Neural Processing Unit (NPU)\cite{npu_evolution}. Moreover, advancements in AI software, such as quantization, have reduced both the size of models and the computational power required to run them, making on-device deployment more feasible and efficient. This form of deployment offers key advantages such as improved data privacy, since data is processed directly on the device, and reduced latency, as inference no longer depends on an internet connection\cite{latency_privacy}.
However, while on-device AI deployment is becoming more popular, its security implications are still often misunderstood, especially when considering how AI differs from traditional software. This gap in understanding can lead to serious vulnerabilities particularly when AI is used in security applications such as content filtering or spam detection.

\begin{figure}
    \centering
    \includegraphics[width=0.4\linewidth]{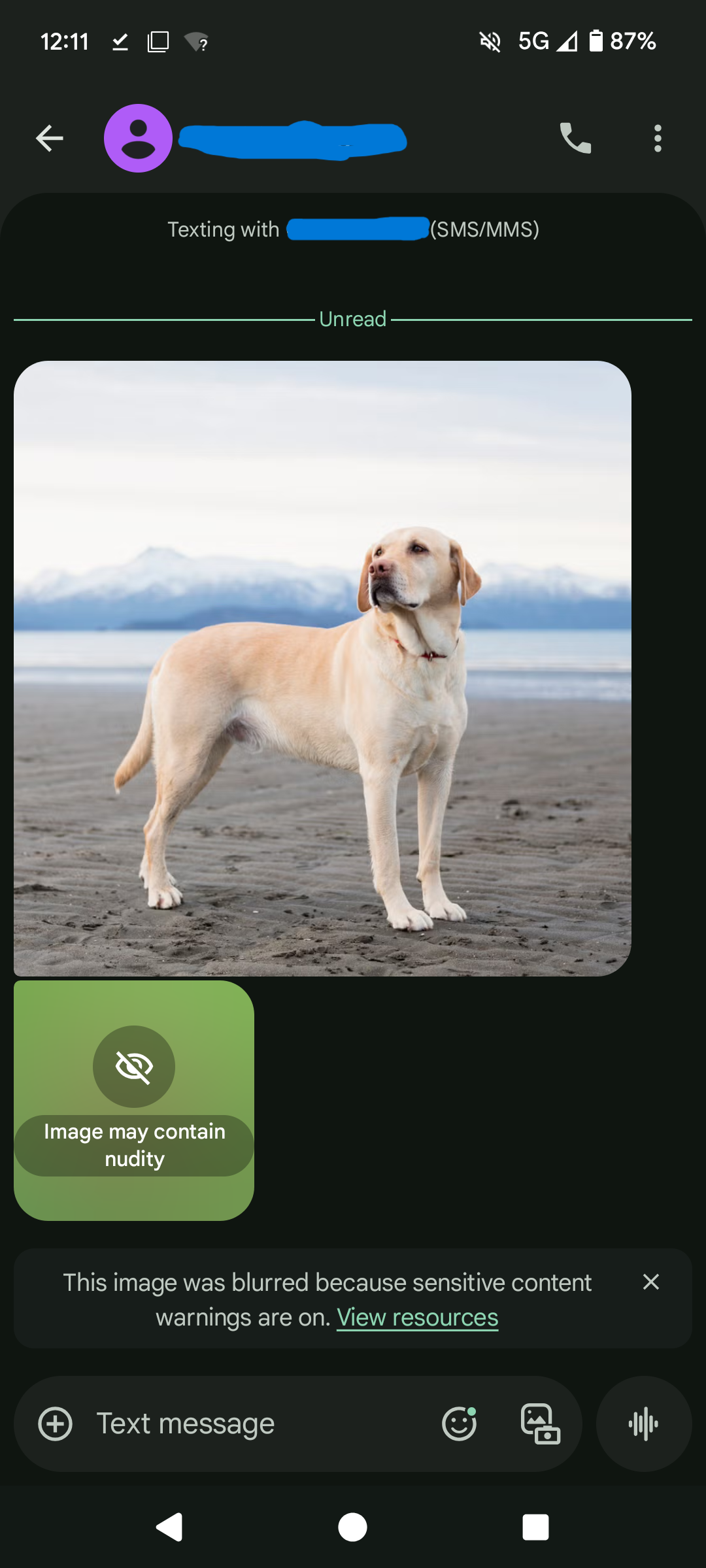}
    \caption{User interface of the Google Messages application with SafetyCore enabled. When nudity is detected by the AI model, the image is blurred and a warning message is displayed to the user.}
    \label{fig:safetycore_app}
\end{figure}

In this article, we explore the security risks associated with on-device AI deployment through the lens of a real-world case study: the exploitation of the SafetyCore application\cite{safetycore}.
SafetyCore is an Android Google system service introduced in November 2024. It provides a service used by other Android applications: The classification of sensitive or problematic content, such as nudity in images. The content detection is performed using an AI based algorithm that is locally embedded on the device for privacy-preserving reasons. This means that the user data is not sent to a remote server, but is kept on the device.
Currently, SafetyCore is only used by the Google Messages application for content moderation. However, other applications, such as WhatsApp\cite{WhatsApp_integration}, are expected to adopt it in the near future.
The AI model used by SafetyCore takes an image as input and predicts whether it contains sensitive content or not. If such content is detected, the image is automatically blurred, and a warning message is displayed to inform the user that the image may contain unwanted characteristics, such as nudity.
An example of the application behavior is shown in Figure \ref{fig:safetycore_app}.

SafetyCore relies on the AI model's ability to make accurate predictions based on pixel values in the image. 
Starting with the extraction of the embedded model, we demonstrate how adversaries can manipulate images to cause misclassifications, thereby rendering the protection mechanism ineffective.
The objectives of this article are twofold:
\begin{itemize}
    \item Explaining the specific risks of deploying AI models on-device, especially for readers without a background in AI.
    \item Providing a practical guide to extracting and exploiting these models in a real-world setting.
\end{itemize}
Each aspect of our analysis is illustrated using the SafetyCore attack case study. 
This attack has been performed on a Google Pixel~6, running Android~15 (build \texttt{BP1A.250305.019}) with SafetyCore (\texttt{com.google.android.safetycore}) version \texttt{1.0.757930370}.

We begin this article by examining the specific methods for reverse-engineering an AI model and discuss what makes AI models fundamentally different from traditional software. We then describe the pre-processing and conversion steps necessary to turn an extracted model into a targeted object. Finally, we explore intrinsic vulnerabilities of AI models and demonstrate how they can be exploited through AI based attacks.

\section{The Reverse Engineering Challenge}
This Section explores what makes AI models fundamentally different from standard code, and why traditional software protections are often insufficient to secure them from reverse-engineering.

\subsection{What is Inside an AI Model?}
\begin{figure}
    \centering
    \includegraphics[width=0.4\linewidth]{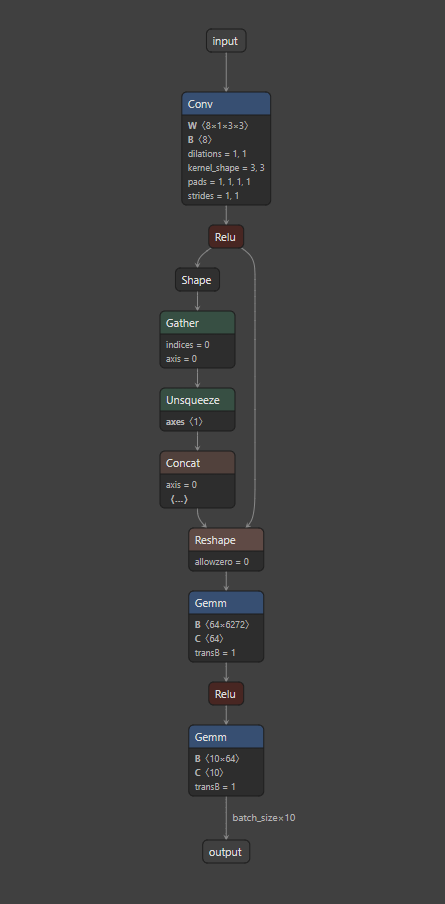}
    \caption{Example of a toy AI model (in ONNX format). Each node of the graph corresponds to a specific layer. The first node is a convolution layer that contains learned parameters $W,b$ and hyperparameters $dilation,kernelShape,pads,stride$.}
    \label{fig:aimodel}
\end{figure}

The goal of an AI model is to perform a given task (prediction, generation) on data never seen before.
It is defined by an architecture composed of layers, hyperparameters and learned parameters. Layers represent mathematical operations, often linear transformations such as matrix multiplications. Each layer has hyperparameters, whose values are fixed before training and remain unchanged.
In contrast, the learned parameters, such as weights and biases, are updated throughout the training process so that the model can perform its task on new data.
An example of a toy AI model is provided in Figure \ref{fig:aimodel}.
The lifecycle of an AI model can be divided into two phases: training and inference. Training is the step where the learned parameters are updated, i.e. the network learns to adapt itself to new data. The inference step is when the model is used to perform some prediction on unseen data. At this stage, the learned parameters are fixed and no longer updated.

%Parameters are constant tensors learned during the training phase of the AI conception.

%These information are typically stored in a model file, which serialize both the sequence of operations and their corresponding parameters. At runtime, an inference engine reads and parses this file. The operations are executed using inputs and parameters and a result is produced: the model prediction.

% Add an illustration of an AI model

\subsection{Why AI Models are Different from Classical Software?}
Software is typically made of code that is compiled and deployed. This means that only the hardware instructions are present on the deployment target.
\textbf{On the other hand, an AI model is usually stored in a file}.
This file does not directly contain the hardware instructions, as traditional software does, but rather a serialized version of the algorithm. This serialized file defines the layer operations as well as the learned parameters needed to produce the model output. The specific implementation of those operations is handled by a separate inference engine. To parse and run a model, the inference engine associated with the model is required.

The operations used during inference are implemented by the inference engine. Thus, a model can only use a limited set of standardized layers. \textbf{When an AI model is run, the executed operations came from the same finite set of layers, regardless of the specific model architecture or parameters values.} The inference engine is contained in a compiled library and defines how the model is serialized. \textbf{A limited set of inference engines are often used to deploy AI models, and most of them are open-source}. This is because layer implementations are highly optimized for specific hardware, making it inefficient and unnecessary for each company developing AI models to re-implement them from scratch for all standard hardwares\cite{not_from_scratch}.

While this standardization is beneficial for performance and cross-platform compatibility, it raises significant challenges regarding intellectual property protection.

\subsection{Static Extraction} \label{sec:static_attack}
The first method to reverse engineer a model is to perform a static analysis (i.e., analysis without running the application) of the application to locate and extract AI models. Since serialization depends on the specific library used, it is important to understand the different libraries and their formats.
The major inference engines used when deploying an AI model on-device include LiteRT\cite{lite_rt} (formerly TensorFlow Lite), ONNX\cite{onnx}, and PyTorch (through TorchScript\cite{torchscript} and ExecuTorch\cite{executorch} formats).
\textbf{Each of these engines reads AI models from a file that has distinctive identifiable characteristics}. This makes it relatively easy to locate such files within a software package, enabling static file analysis.
\autoref{fig:magic_numbers} lists the different characteristics that can be searched for in the files of an application to locate AI models.

\begin{table*}[t]
\centering
\caption{Characteristics to identify AI models used by major inference engines.}
\label{fig:magic_numbers}
\begin{tabularx}{\linewidth}{|l|X|}
    \hline
    \textbf{Inference engine} & \textbf{Characteristic} \\
    \hline
    LiteRT / TensorFlow Lite & FlatBuffers file identifier* \texttt{TFL3} (ASCII encoded) \\
    \hline
    ONNX & Protobuf containing the graph. Each layer type starts with \texttt{onnx::} \\
    \hline
    TorchScript &
    ZIP archive (file signature \texttt{PK\textbackslash x03\textbackslash x04}) containing:
    % Using enumitem for more compact lists within the table
    \begin{itemize}[nosep, leftmargin=*, topsep=2pt, partopsep=0pt]
        \item The directories:
            \begin{itemize}[nosep, leftmargin=*]
                \item \texttt{code}
                \item \texttt{data}
            \end{itemize}
        \item The files:
            \begin{itemize}[nosep, leftmargin=*]
                \item \texttt{data.pkl}
                \item \texttt{constants.pkl}
            \end{itemize}
    \end{itemize} \\
    \hline
    ExecuTorch & FlatBuffers file identifier* \texttt{ET??} followed by \texttt{eh??} (ASCII encoded), where \texttt{?} is a digit. \\
    \hline
\end{tabularx}
\par\medskip
\small *The \href{https://flatbuffers.dev/schema/#file-identification-and-extension}{FlatBuffers file identifier} is a field in the FlatBuffers serialization format.
\end{table*}

\paragraph{The SafetyCore case}
In the case of SafetyCore, static analysis of the service’s downloaded files revealed a TensorFlow Lite model. The presence of the ASCII-encoded magic value \texttt{TFL3} at byte offset 4 led to a quick identification of the model file.

\subsection{Dynamic Extraction, or Why Encryption Is Not Enough}
In some cases, the AI model in plain-text form never touches persistent storage. For instance, when it is downloaded (remote loading) and loaded directly into memory for each inference, or stored only in encrypted form (model encryption).

\begin{itemize}
    \item \textbf{Remote loading} avoids static interception of the model file as it is never stored in persistent storage.
    \item \textbf{Model encryption} effectively hides the file structure, making it impossible to locate using known characteristics of the model while performing a static analysis.
\end{itemize}

In such situations, dynamic analysis can be used to intercept the serialized model at runtime, capturing it while it is being loaded into the inference engine. This was not the case for SafetyCore, where the model was recovered through static analysis. In practice, encryption and remote loading can be bypassed, and the model extracted using dynamic analysis. While having privileged access on the running device, instrumentation tools such as Frida\cite{frida} can be used to hook the model loading functions and exfiltrate the model during execution. \textbf{Since most inference engines are open source, it is relatively easy to identify and hook the model loading function, even if it is not directly exposed by the library.}

\section{AI Model Refinement}
After this first reverse-engineering step, an AI model often requires refinement before it can be effectively exploited.
\subsection{Convert to the Right Format} \label{sec:right_format}
\textbf{Extracted AI models are often not immediately usable by attackers, because they are deployed in formats that do not support gradient computation} (Additional details about gradients are provided in Section \ref{sec:vulnerabilities}).
PyTorch is the most widely used framework for attacking AI models. In contrast, formats such as TFLite and ONNX do not natively allow gradient computation, making them not suitable for direct exploitation.
Therefore, converting the extracted model to PyTorch is typically a necessary step. This conversion can often be achieved using available tools, either directly or through a combination of intermediate formats.
In Table~\ref{tab:conversion_tools} is presented common AI model formats and the corresponding tools used to convert them into PyTorch.
\begin{table}[h]
\centering
\begin{tabular}{|l|l|}
\hline
\textbf{Original Format} & \textbf{Conversion Tool(s)} \\
\hline
ONNX & \texttt{onnx2pytorch\cite{onnx2torch}} \\
TFLite & (\texttt{tf2onnx} + \texttt{onnx2pytorch}) via REOM\cite{REOM} \\
TorchScript & Natively exploitable \\
ExecuTorch & Not currently supported* \\
\hline
\end{tabular}
% Add ref to REOM
\caption{Conversion tools for enabling PyTorch based attacks on extracted AI models.}
\small *Primarily due to the novelty of the framework compared to more mature alternatives such as TFLite.
\label{tab:conversion_tools}
\end{table}
\subsection{The Quantization Problem}
Quantization refers to the process of converting the learned parameters of an AI model from high-precision floating-point (typically \textit{float32}) to lower-precision formats such as 8-bit integers (typically \textit{int8})\cite{quantization}. This transformation reduces both memory usage and computational cost, making it particularly suitable for on-device deployment, where hardware resources are limited\cite{quantization}.

The most widely used approach is \textit{affine quantization}. This method relies on two quantization parameters:
\begin{itemize}
    \item a scale factor  $s \in \mathbb{R}^+$.
    \item a zero point $z \in \mathbb{Z}$.
\end{itemize}
These parameters are used both to:
\begin{itemize}
    \item Convert (quantize) the original \textit{float32} parameters to integers.
    \item Compute operations directly in the quantized domain (integer domain).
\end{itemize}
Given a real-valued parameter $w \in \mathbb{R}$, its quantized representation $w_q \in \mathbb{Z}_{[\alpha_q,\beta_q]}$ is computed as:
% better explaiin why int8 
\begin{equation*}
     w_q = \text{clip}\Big( \text{round}\big(\frac{1}{s} w + z\big), \alpha_q, \beta_q \Big)
\end{equation*}

Since the model no longer operates on differentiable \textit{float32} parameters, standard gradient-based techniques cannot be directly applied to a quantized model. However, it is important to note that \textbf{quantization does not act as a security mechanism}. The combination of quantized parameters and their associated scale and zero point is sufficient to reconstruct an approximation of the original parameters.
For $w \in \mathbb{R}$ we have:

\begin{equation} \label{eq:dequantize}
    w \approx \text{dequantize}(w_q, s, z) = s \cdot (w_q - z)
\end{equation}
Using this equation, an attacker can construct a proxy model, i.e., a model that approximates the behavior of the original one.
This proxy model is fully differentiable and can be attacked using standard gradient-based methods.

\paragraph{The SafetyCore case}
In the case of the SafetyCore application, the target model was provided in the TFLite format. As shown in Table~\ref{tab:conversion_tools}, REOM\cite{REOM} allows the conversion of a TFLite model to PyTorch by leveraging a combination of two intermediate conversion tools. Additionally, REOM integrates a quantization module that applies Equation~\ref{eq:dequantize} to recover \textit{float32} parameters from \textit{int8} parameters, enabling the construction of the proxy model.
The tool successfully generated a \textit{float32} proxy model that could be subjected to further attacks\footnote{For our proxy model, we did not add additional layers to simulate quantization errors, as we observed no significant differences between the quantized and the reconstructed model.}. 

\section{Exploiting AI Models}
After transforming a model into a usable artifact, we analyze the vulnerabilities of AI systems and the unique security challenges they pose. We then present how to exploit these vulnerabilities through adversarial examples. Finally, we discuss additional attacks that are relevant once an AI model is extracted.

\subsection{Intrinsic Vulnerability of AI Models} \label{sec:vulnerabilities}
The intrinsic vulnerability of AI relies on three intricate problems:
\paragraph{Gradient manipulation}
A neural network is a highly complex function that takes an input and, using a set of parameters, produces an output. These parameters must be learned in order for the model to generate meaningful results.
To learn these parameters, we define an auxiliary function, the loss function, which tells us how much the model is wrong in its prediction. For instance, in an image classification task, the loss function could quantify how inaccurately the model distinguishes between images of cats and dogs. The goal is typically to minimize this loss function.
Training the model involves updating its parameters to reduce the loss, using a dataset of input/output pairs (known as the training set). This process is often performed using an algorithm called gradient descent, which iteratively adjusts the parameters in the direction that reduces the loss.
The gradient is a mathematical object that indicates how to change the parameters to minimize the loss.  

During the training phase, the gradients are computed with respect to the model parameters to minimize the loss.
\textbf{Once the model is trained, however, an attacker can instead compute gradients with respect to the input, this time to \emph{maximize} the loss i.e. make the model's prediction as wrong as possible.
In this setting, the gradient reveals how the input should be perturbed to mislead the model.}
Because neural networks are highly complex and operate in high-dimensional input spaces, these perturbations can be crafted so that they remain imperceptible to humans, making them particularly dangerous.

\paragraph{The black-box problem}
Despite their remarkable performance, AI models are black-boxes in the sense that the decision-making process is not understandable by humans\cite{XAI}. In other words, given a particular input, we often cannot understand why the model produces a specific output\cite{output_why}.
Although the field of explainable AI (XAI) has made significant progress, it remains difficult to predict how a model will behave on unseen or slightly altered inputs. This inherent opacity creates ``gray areas" of unpredictable or unintuitive behavior that are exploitable. \textbf{These unpredictable behaviors are not easily identifiable or interpretable by human observers, making them ideal entry points for adversarial manipulation.}
\paragraph{Features correlation}
AI models typically rely on statistical correlations in the training data rather than causal relationships. This distinction is critical: a model might learn that ``A" often co-occurs with ``B," without grasping whether ``A" causes ``B."
\textbf{This reliance on correlation rather than causation contributes to unexpected and or unintelligible model behavior that can bypass human judgment.}

\subsection{Exploiting These Vulnerabilities}
The vulnerabilities presented above can be exploited in multiple ways across the AI lifecycle.
In this Section, we focus on concrete attack strategies that target on-device AI models when having access to the architecture and parameters.

\paragraph{Inferring the loss function}
Many attacks rely on gradient computations, which not only require knowledge of the model’s architecture and parameters but also defining a suitable loss function.
The choice of this loss function is driven by the identification of the task the model is solving, for example a binary classification task.
\textbf{A common attack strategy is to identify the loss that was used for model training, and use the opposite for attacking (in order to maximize it).
While this information is typically unavailable after deployment, attackers can often infer it through careful analysis of the architecture and model behavior.}

This process typically involves the following steps:

\begin{enumerate}
% Use netron !!!
\item \textbf{Architecture probing}: By analyzing the metadata (e.g., input/output shapes, presence of specific layers such as convolutions or residual blocks, and even embedded strings in the model file), one can make educated guesses about the model architecture and its intended task.
For some model formats, such as TFLite or ONNX, the overall architecture can be visualized using a visualization tool like Netron\footnote{\url{http://netron.app}}.
\item \textbf{I/O probing}: By feeding the model with various sample inputs and observing the output responses, one can understand the appropriate input format and the semantics of the outputs (classification scores, images, heatmaps etc…).
\item \textbf{Output layer inspection}: The final activation function often reveals the nature of the learning problem. A softmax activation suggests a classification task, and a sigmoid (logistic) activation a multi-label classification task. A linear output usually indicates regression.
\end{enumerate}
\paragraph{The SafetyCore case}
In the case of the SafetyCore, the input tensor shape is $1 \times 224 \times 224 \times 3$, which strongly suggest image data. The architecture includes residual connections common in ResNet architectures\cite{resnet} for image classification. The output shape is $1 \times 4$, and when probing the model with explicit versus non-explicit images, we observe that explicit content causes some output values to exceed $0.5$, while benign content remains below this threshold. This, along with the presence of a sigmoid activation function before the output, suggests that the model is solving a multi-label classification problem, and was trained using a binary cross-entropy loss. This inferred loss enables the attacker to compute gradients for further manipulations.

\subsubsection{Adversarial Examples}
\textbf{The philosophy behind adversarial examples involves defining a desired criterion on the model's output through a loss function, and then using model gradients to find an input modification that satisfy this criterion.}
This attack relies mostly on the properties of high dimensional spaces and the nature of the functions learned by deep learning models. In these spaces, a tiny perturbation, when applied in a specific direction (using gradient), can often cause a significant change in the model's output.

Generally, the input change is sought to be imperceptible and to maximize the model's output change\cite{fsgm}.
For classifiers this could mean altering a picture of a panda so that the model confidently predict a gibbon\cite{fsgm}.
There are two main types of adversarial attacks:
\begin{itemize}
    \item \textbf{Untargeted attacks:} The goal is to mislead the model, regardless of what that incorrect output is.
    \item \textbf{Targeted attacks:} The goal is to produce a pre-determined output.  The attacker doesn't just want the model to be wrong, he wants it to produce a particular output.
\end{itemize}

For generating these examples, a diverse range of attack algorithms has been developed, from the fast gradient sign method (FGSM)\cite{fsgm} to more iterative and powerful methods like Projected Gradient Descent (PGD)\cite{pgd}, each with different trade-offs in terms of computational cost and effectiveness\cite{adversarial_examples_survey}.

It is important to note that the effectiveness of adversarial examples depends on the data modality of the input. They are particularly effective on continuous data, such as images and audio, where small perturbations can be applied directly to the input using gradient-based methods\cite{pgd}. In contrast, generating adversarial examples for discrete data (like text) is more challenging, as it is difficult to generate discrete changes in a meaningful way using gradients\cite{adversarial_example_text}.
However, many text models include both discrete and continuous components. In such cases, it is often possible to generate adversarial perturbations in the continuous space and then extrapolate them back to the discrete space\cite{adversarial_example_text}.

Adversarial examples are extremely powerful in practice.
Because of the high dimensionality of the input space, it is not feasible to simply ``patch" a given adversarial example by specifically instructing the model to ignore it during training.  Doing so leaves the model vulnerable to countless regenerated variants that came from the same region of the input space.
A common defense strategy is adversarial training\cite{adversarial_training}, which involves incorporating multiple adversarial examples in the model training.
While it offers a potential defense, it remains difficult to fully mitigate the threat, as this approach often involves a trade-off with model performance\cite{adversarial_training}.

\paragraph{The Safetycore case}
For SafetyCore, we implemented a \textbf{Projected Gradient Descent (PGD)} attack, a widely used method known for its effectiveness in white-box setting (e.g when access to the model parameters).

This allows two types of attacks:

\begin{enumerate}
    \item \textbf{False Positive (Enable Blurring):}
       You can start from a non-explicit image and generate a small, imperceptible perturbations to make the model predicting it as explicit. As a result, SafetyCore will apply blurring to an image that should not be blurred.
    \item \textbf{False Negative (Bypass Blurring)}
        Yo can also start from an explicit image. This image will contains a small perturbation that prevent the model from recognizing it as explicit. Consequently, SafetyCore will not apply any blurring to it, effectively bypassing the protection.
\end{enumerate}

In Listing \ref{pgd:code} a PyTorch implementation of the attack script is provided. 
This script is intended to be simple and as generic as possible and can be reused on other models as long as the input data is continuous, and a loss function can be chosen.
The most important parameters of this script are $\epsilon$ and $num\_iter$.
\begin{itemize}
    \item $\epsilon$ control the maximum change per pixel that is expected for the perturbed input. Higher values mean higher pixels variations and then more visible perturbations.
You can gradually increase $\epsilon$ until you find a sample with the expected model output.
    \item $num\_iter$ control the number of iteration (number of gradient steps) that are taken during the attack. A higher number of iterations allows finding a more effective adversarial example in terms of loss function maximization.
\end{itemize}
Before running the attack script, the input images are resized to $1 \times 224 \times 224 \times 3$ to match the model’s input dimensions. The resulting adversarial examples have the same size and are saved in a .png format.
It is important to use an image format that does not apply compression in order to preserve the adversarial perturbation (avoiding JPEG, which would remove part of the added noise).

\begin{figure}
    \centering
    \includegraphics[width=0.5\linewidth]{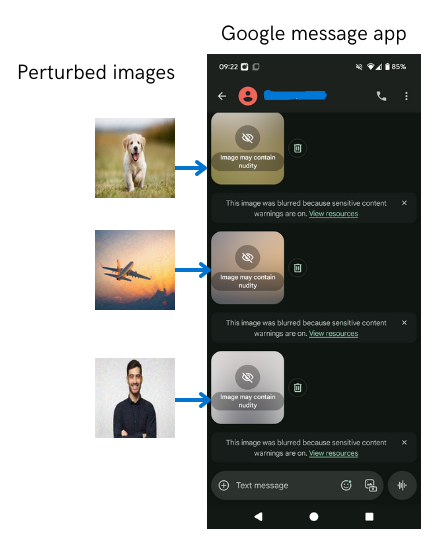}
    \caption{Adversarial attack on SafetyCore. On the left, are provided the perturbed images obtained after a PGD attack, and on the right what appears in the Google Message application when sent to an Android device.}
    \label{fig:safetycore_perturbed}
\end{figure}

In Figure \ref{fig:safetycore_perturbed}, we show three benign images that have been perturbed using PGD (Enable Blurring case). When passed through the originally extracted model, these adversarial images are misclassified as explicit content. In Figure \ref{fig:safetycore_perturbed}, we also present a screenshot from the Google Messages app showing how these images appear to users: all are blurred with a warning about potential explicit content.
You can imagine the same scenario with explicit images that will not be blurred in the application (Bypass Blurring). For ethical reasons, this case is not illustrated in the article. Executing an ``enable blurring attack" or a ``bypass blurring attack" follows the same methodology, allowing the adversary to choose either at will.

\textbf{The implications of this attack are severe: since the same AI model is shared across all Android devices, it is possible to fully bypass the filtering capabilities, regardless of the input. 
This attacks takes less than 30 lines of code.}

\subsubsection{Additional Relevant Attacks}
Even if it is very powerful, adversarial examples generation is not the only relevant attack that can be performed after extracting an AI model.
In this Section we presented two additional types of methods that can be applied.

\paragraph{Model inversion}
Model inversion attacks aim to reconstruct representations of the training data by using the AI model itself\cite{model_inversion}. Typically, these attacks exploit model gradients to iteratively optimize an input that maximizes the model’s confidence\cite{model_inversion}. This technique is most commonly applied to classification tasks, where the goal is to generate inputs that are representative of a target class.
The consequences can be severe, often resulting in the leakage of private data for example, reconstructing faces that were used to train a facial recognition model.
Additionally, model inversion can provide insights into the task the model was trained on. For instance, if the reconstructed inputs resemble airplanes and the model architecture indicates a classification task, one can reasonably infer that the model is likely classifying different types of planes from images.

In the context of SafetyCore, we attempted a model inversion attack. However, the optimization process quickly collapsed. This failure can be attributed to two main factors. First, the training data for each class is probably highly diverse, making it difficult for the model to converge toward a shared representation. Second, the task is a multi-class classification problem, which further complicates the inversion process, as each class may share overlapping features with others.

\paragraph{Backdoor Attacks}
Backdoor attacks involve poisoning the training data with specially crafted samples that cause the model to learn spurious correlations between a trigger pattern and a specific output\cite{backdoor}. When the attacker later inputs a sample containing the same trigger, the model exhibits the intended behavior\cite{backdoor}.

This type of attack exploits two vulnerabilities described in Section \ref{sec:vulnerabilities}:
\begin{itemize}
    \item The model is learning arbitrary associations (like small patch for images) that have no causal relation to the task.
    \item The black-box nature of AI means that such malicious correlations are difficult to interpret or detect after training. 
\end{itemize}
In the case of SafetyCore, this type of attack may be unnecessary, as adversarial examples can achieve the same effect without requiring the model to be retrained. Moreover, there is no indication that the submitted images will be used for future training. In fact, since the model runs entirely on-device, retraining seems unlikely. However, unlike adversarial examples, which may not transfer to new model versions, backdoor examples are more likely to persist. Indeed, as backdoors are difficult to detect in training data, such examples could remain in future training sets, potentially preserving the backdoor across versions.

\begin{lstlisting}[language=Python, caption=Small generic Python code for generating adversarial examples with PGD, label={pgd:code}]
import torch 
from torch import nn

def pgd_attack(
    model, inputs, targets, epsilon=0.01, alpha=0.005, num_iter=100,
    loss_fn=None, random_start=True, clip_min=0.0, clip_max=1.0):
    """
    Projected Gradient Descent (PGD) attack on a PyTorch model.

    Parameters:
    -----------
    model : torch.nn.Module
        The neural network to attack.
    inputs : torch.Tensor
        Original input images or continuous data to perturb.
        TODO: Replace by your own input
    targets : torch.Tensor
        Ground truth labels corresponding to the inputs.
    epsilon : float
        Maximum perturbation allowed (L-infinity norm).
    alpha : float
        Step size for each iteration.
    num_iter : int
        Number of attack iterations.
    loss_fn : callable, optional
        Loss function to maximize (defaults to BCELoss if None).
        TODO: Replace by your own loss function
    random_start : bool
        If True, start from a random point within the epsilon-ball around the input.
    clip_min : float
        Minimum allowed value for perturbed inputs.
    clip_max : float
        Maximum allowed value for perturbed inputs.

    Returns:
    --------
    torch.Tensor
        Adversarially perturbed inputs.
    """
    model.eval()
    original_inputs = inputs.clone().detach()

    if random_start:
        # Start from a random point within the epsilon-ball
        adv_inputs = original_inputs + torch.empty_like(inputs).uniform_(-epsilon, epsilon)
        adv_inputs = torch.clamp(adv_inputs, clip_min, clip_max)
    else:
        adv_inputs = original_inputs.clone().detach()

    if loss_fn is None:
        # Loss function used for the attack
        # TODO: Replace by your own loss function
        loss_fn = nn.BCELoss()

    for _ in range(num_iter):
        adv_inputs.requires_grad_(True)
        outputs = model(adv_inputs)
        loss = loss_fn(outputs, targets)

        model.zero_grad()
        loss.backward()
        grad_sign = adv_inputs.grad.detach().sign()

        adv_inputs = adv_inputs + alpha * grad_sign
        # Project back to the epsilon-ball and clip to valid range
        perturbation = torch.clamp(adv_inputs - original_inputs, min=-epsilon, max=epsilon)
        adv_inputs = torch.clamp(original_inputs + perturbation, clip_min, clip_max).detach()

    return adv_inputs

\end{lstlisting}

\section{Conclusion}
Security should serve as a cornerstone for building trust in AI systems. In this paper, we explored the risks of deploying AI models on-device through the lens of the SafetyCore application. Our work demonstrates that once adversaries gain access to the model, it can be compromised with relative ease and rendered ineffective, raising important concerns for the security of on-device AI based applications. While on-device AI enables countless use cases, its specific security challenges are still overlooked, even by major players in the field as illustrated by the SafetyCore example. This work acts as a foundation for understanding and running attacks on on-device AI models, and can be extended to a wide range of applications. In future work, we plan to extend our methodology to more data modalities and use-cases.

% This process can be apply to other android apps

\bibliographystyle{unsrtnat}
\bibliography{references}  %%% Uncomment this line and comment out the ``thebibliography'' section below to use the external .bib file (using bibtex) .

\end{document}